Book chapter

# Memory and Planning in Brains and Machines: Multiscale Predictive Representations

Ida Momennejad
Microsoft Research

**Abstract**

Memory is inherently entangled with prediction and planning. Flexible behavior in biological and artificial agents depends on the interplay of learning from the past and predicting the future in ever-changing environments. This chapter reviews computational, behavioral, and neural evidence suggesting these processes rely on learning the relational structure of experiences, known as cognitive maps, and draws two key takeaways. First, that these memory structures are organized as multiscale, compact predictive representations in hippocampal and prefrontal cortex (PFC) hierarchies. Second, we argue that such predictive memory structures are crucial to the complementary functions of the hippocampus and PFC: enabling a recall of detailed and coherent past episodes, and generalizing experiences at varying scales for efficient prediction and planning. These insights advance our understanding of the brain's memory and planning mechanisms, and hold significant implications for advancing artificial intelligence systems.

Imagine planning an overseas trip. You could consider the plan in exhaustive detail, down to which seat you prefer, what you will pack, your commute plan, which socks to wear. Or you could consider the most crucial steps, thinking about it at a high level by jumping over many steps and merely considering key steps such as getting a plane ticket and booking a hotel. In this imaginary scenario you easily use representations of the world that are organized in your memory at multiple scales. This example illustrates the central thesis of this chapter. The ability to move back and forth across representations of the world, to remember the past and imagine and plan the future at different scales of abstraction, is crucial to the structure of memory and its use in prediction and planning.

What should these memory structures look like to enable planning and how do we study them? This chapter reviews evidence showing that the learnt structures of events, such as the spatial structure of an environment (Figure 1) or the relational structure of a social network, are organized in memory as predictive representations that are multistep and multiscale. That is, brain regions can organize the same memories with different predictive horizons or scales (Figure 1). The scale varies depending on the task at hand or the level of abstraction involved. Much of this evidence is derived from experimental areas where the roles of memory in planning are particularly evident, including spatial navigation and non-spatial associative inference.

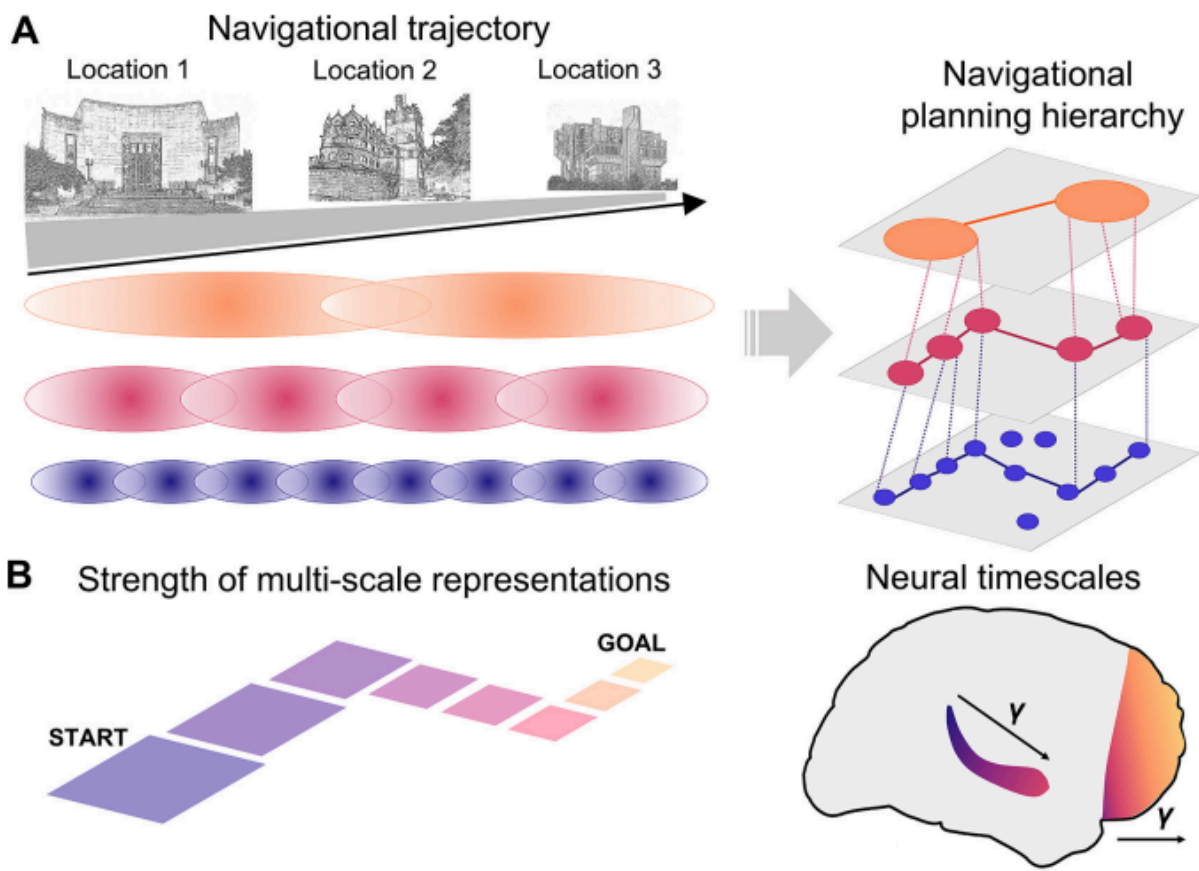

**Figure 1. Multi-scale predictive representations in navigation and planning.**
(A) A navigational trajectory can be represented with varying granularity: e.g., step by step (purple), block by block (pink), or subgoal to subgoal (orang). (B)

During planning, the scope or horizon "visible" each scale covers can vary from a step to the goal. Here review evidence that these gradients of scales are reflected in representational hierarchies in the hippocampus and the prefrontal cortex, and are called on flexibly in navigational memory and planning.

*What is a cognitive map?* A cognitive map is a representation of latent relational structures that underlies a task or environment. It is how our knowledge or memories are organized in order to facilitate efficient retrieval, planning, reasoning, and inference in biological and artificial agents (Tolman 1948; Kumaran and Maguire 2005; Epstein et al. 2017; Behrens et al. 2018; Ida Momennejad 2020; Brunec and Momennejad 2021).

The concept originated from Tolman's latent learning experiments, demonstrating rodents' ability to learn the latent structure of a maze without rewards (Tolman 1948). This challenged the dominant behaviorist dogma of the time that learning only occurs with reinforcement; and paved the way for a cognitivist revolution. Decades later, discoveries of hippocampal place cells (J. O'Keefe and Dostrovsky 1971; J. O'Keefe 1976; John O'Keefe and Nadel 1978) and entorhinal cortex grid cells (Fyhn et al. 2004; Hafting et al. 2005; Moser, Kropff, and Moser 2008), together referred to as "the brain's GPS," further substantiated cognitive maps and earned the 2014 Nobel Prize ("The Nobel Prize in Physiology or Medicine 2014" n.d.).

Cognitive maps have since been studied behaviorally, computationally, and neurally. More recent studies suggest that multi-step, multi-scale, and compressed neural representations are crucial for inference in both memory and planning (Behrens et al. 2018). Over the past decades, a number of Reinforcement Learning (RL) and deep neural network models have been proposed to capture the computations involved in cognitive maps and planning, especially in the hippocampus and the prefrontal cortex of humans, rodents, bats, monkeys, and birds (Epstein et al. 2017; Behrens et al. 2018; Brunec and Momennejad 2021; Pudhiyidath et al. 2022).

How does your brain organize cognitive maps? *How* is the organization of memory (the past) connected to prediction and planning (the future)? One possibility is that the brain may store and unroll a step-by-step map of the environment for planning. This is similar to the idea of model-based reinforcement learning (MBRL) (Sutton and Barto 2018), which we will discuss in the following section. Note that MBRL is indeed a method that allows for multi-step planning, and the options framework allows for multi-step compression (Sutton et al. 2023; Sutton, Precup, and Singh 1999). However, the point of discussion here is whether the *representation* of the environment, or the map of its states, is stored as multi-step compressed representations, or as one-step tuples that can be unfolded for planning.

Another possibility is that the representational structure of memory is both multistep and multiscale. This allows for "jumps" over multiple steps in memory, i.e., how many steps a compressed representation "jumps" over. Here the horizon or "scale" of these jumps rely on compression and the predictive nature of the representation at a given scale. Similarly, computational and empirical studies of cognitive maps suggest that they are organized as predictive representations that are also multiscale, organized with different predictive scales, or horizons (Figure 1). While there are various approaches to temporal abstraction, one idea highlighted in this chapter involves the successor representation (SR) in reinforcement learning, proposed by Dayan, and variations that combine SR with replay and propose multi-scale successor representations (Dayan 1993; I. Momennejad et al. 2017; Ida Momennejad 2020; Machado et al. 2023). We will discuss this in more detail in Sections 1 and 2.

Accumulating evidence over the past decades is consistent with the idea that such predictive representations may govern human behavior in episodic memory tasks (Gershman et al. 2012), planning and decision making (I. Momennejad et al. 2017; Russek et al. 2017), and further, that these compressed representations may support cognitive maps (Behrens et al. 2018) in the rodent hippocampus and entorhinal cortex (Stachenfeld, Botvinick, and Gershman 2017; Jesse P. Geerts et al. 2020; de Cothi et al. 2022).

Consistently, a recent human fMRI study showed that naturalistic VR navigation relies on multiscale predictive representations in prefrontal and hippocampal hierarchies (Brunec and Momennejad 2021). This study showed that weighted representational similarity at longer scales are associated with goal-directed naturalistic navigation. Consistently, another study suggests the brain uses hierarchical representations in anticipation of events that are multiple steps away (Tarder-Stoll, Baldassano, and Aly 2023). A study of complementary task representations in hippocampus and prefrontal cortex showed that task abstractions in medial prefrontal cortex simultaneously represent behavior over different temporal scales (Samborska et al. 2022). More evidence discussed in Section 3 onward.

In this chapter I will first establish how temporal abstraction connects the past and the future, then introduce the main computational methodology (reinforcement learning, RL) for framing this connection in terms of predictive representations at multiple scales, and then discuss a number of behavioral, neural, and computational studies that support the computational idea. I will end by discussing why these approaches can inspire novel ways to evaluate and augment planning and navigation abilities of generative AI.

## 1. Temporal Abstraction: Binding the past and future

Consider how we come to learn representations of the world. One possibility is that as we navigate the world our brains store each event that we experience individually, and later on retrieve them individually. In this framework, planning requires rolling out and evaluating every single event step by step to identify the optimal planning trajectory. This is similar to model-based RL (MBRL, Figure 2).

Let us consider another hypothesis about how brains learn, store, and update representations of the world. As we navigate environments, each event leaves various traces in our memory systems, and these traces last for variable durations, their traces decaying at different rates and scales in different parts of the brain. Over time the trailing traces of events that occurred closer in time overlap, leading to an association among events that are closely associated.

Gradually these overlapping traces are consolidated as associations that, depending on the horizon and decay rate of the trace, lead to temporal clustering of events that occurred within overlapping horizons (Figure 1). This is also a form of compression, or temporal abstraction, which can further lead to the efficient storage of memories, categorization, or the ability to plan. The primary scales or horizons may depend on the size and complexity of the environment in spatial settings, or on the frequency of surprising events more broadly for both spatial and non-spatial associative settings.

Now let us consider how the overlap of trailing traces and the ensuing temporal abstractions serve prediction. Let us say the memory of event A (Figure 1A, location 1) occurred at time t1, and event B (Figure 1A, location 2) at time t3, and event C (Figure 1A, location 3) at time t5. Now consider a part of the brain where the trace of each event lasts for three time steps. As such, each time the memory of event A is triggered by the world, we get a partial activation of event B even when it is not present, because the memory of event A is already bound to the activation of event B so each time A is activated, B is partially activated too.

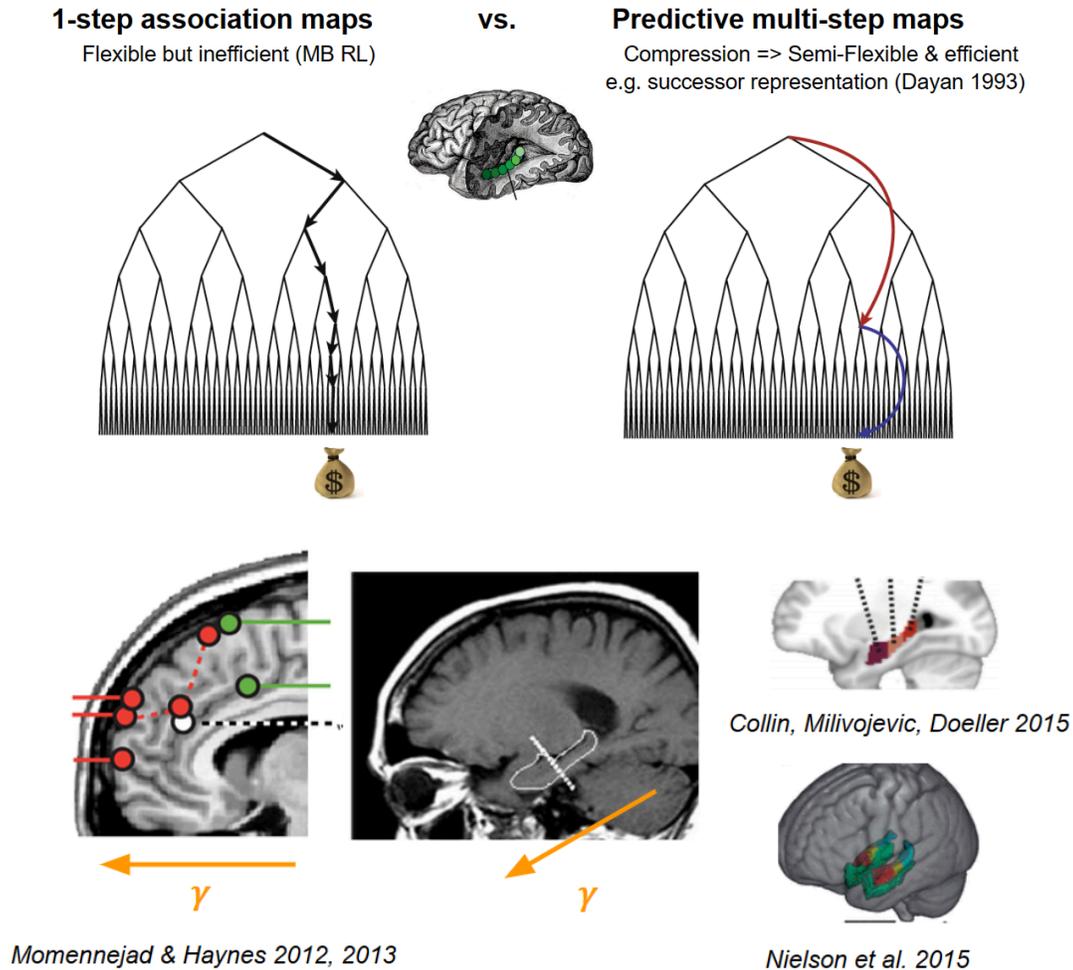

**Figure 2. What do cognitive maps look like?** (top) Two computational hypotheses about the structure of cognitive maps in memory. One possibility is that the brain stores step-by-step (1-step) associations and unrolls them for evaluation at the moment of decision making. This is similar to the transition probabilities in model-based RL (Sutton and Barto 2018). Another possibility is that the brain may store jumpy multi-step associations using *temporal abstraction*, simplifying prediction and planning. Computationally, this is the successor representation (Dayan 1993). (bottom) Evidence from human fMRI suggests that predictive representations may be learned at different scales along the hippocampal and prefrontal hierarchies such that more anterior regions are more likely to represent larger scales, $\gamma$, corresponding to more temporal or spatial abstraction.

As coactivation of successor events occurs over time, an event's activation leads to the partial activation of its successor state in a predictive manner (Figure 1, C).

The strength and scope of the chain of successor event activation depends on two factors. First, it depends on the *distance* of each state from the current one. Second, it depends on the horizon of temporal compression in the part of the brain where these representations are activated.

Evidence from human fMRI (Figure 2) suggests that the scale or horizon of the temporal abstraction increases along the posterior to anterior axis of the hippocampus and entorhinal cortex (Strange et al. 2014; Nielson et al. 2015; Collin, Milivojevic, and Doeller 2015), as well as the rostro-caudal axis of the prefrontal cortex during *goal-directed navigation* (Brunec and Momennejad 2021; Patai and Spiers 2021), *multi-step planning* in sequential and hierarchical problem solving (Botvinick and Weinstein 2014), and *prospective memory* (Paul W. Burgess et al. 2008; I. Momennejad and Haynes 2012; Ida Momennejad and Haynes 2013).

The above mentioned hypotheses can be formalized in the service of computational modeling and quantitative analysis of representations underlying memory, navigation, and planning. Section 2 considers a common approach for such a formalization, the computational framework of reinforcement learning.

2. **The computational approach: Framing memory and predictive representations with reinforcement learning**

Reinforcement learning (RL) is a computational framework that simplifies value-based and goal-directed decision making and planning behavior. The RL framework posits an agent in an environment with a number of states, such that from each non-terminal state $S$ the agent can take action, $a$, to move to the next state $S$' and receive reward $r$ *(Sutton and Barto 2018).*

In classic RL, the agent's goal is often simplified as maximizing expected value in this environment. Most RL algorithms aim to capture how an agent achieves the goal of maximizing value in exploration, learning, and decision making phases. That is, first, how RL captures how the agent takes actions in the environment to explore states, learn the state values or state-state relationships and rewards. Second, RL captures how the agent combines this knowledge to find the policy or sequence of actions, $\pi^* or \pi(a \mid S)$ that maximizes value. This is known as optimal policy, leading to optimal value or V*.

This simple computational framework offers various algorithms for learning representations with abstraction, hierarchy, generalization, and transfer. The most commonly known RL algorithms within the cognitive and neurosciences include model-free RL, model-based RL, the successor representation, and hybrid algorithms combining one or more elements of these algorithms with each other and with offline replay. Let us take a closer look at them.

## 2.1. Model-free RL, model-based RL, and the successor representation

Briefly, model-free agents simply learn to cache the value of each state and action, *Q(S, a)*, in a lookup table or a value function *V(S)*. Importantly this approach combines states, actions, and rewards to cache a scalar expected value without storing state-state relationship, i.e., the probabilities of states leading to one another. Since model-free RL does not store representations of the relational structure of states, it *can not* serve as a candidate for formalizing the structure of cognitive maps and memory. The approach has been fruitful in the study of simpler value-based decision making and "habits" (Gläscher et al. 2010), and in modeling striatal learning in the dual-systems hypothesis (Daw, Niv, and Dayan 2005).

Let us consider an RL algorithm that can serve as models of memory. Model based RL agents (MBRL) capture the relationship between adjacent states, that is, the probability of transition to a state S' given the agent starts in state S and takes action a. This is formalized as *p(S' | S, a)*. As such, MBRL stores the reward for state and action tuples, r*(S, a)*, on the one hand; and the relational structure of the environment, , *p(S' | S, a)*, on the other, i.e., a map of one-step transition probabilities (Figure 2, Table 1).

Unlike model-free RL, the model-based agent does not use a look-up table of cached values to determine the optimal policy (or the sequence of actions that yield the highest value given the goals). Instead, MBRL rolls out, step-by-step, all possible policies iteratively, computing the expected discounted value of each trajectory and picking the policy, or sequence of actions, that maximizes rewards. Updating the value function relies on iterations using the Bellman equation:

$$Equation\ 1.\ V^*(S)\ =\ max_a[R(S,\ a)\ +\ \sum_{S'} p(S'|S,\ a)\ \gamma V^*(S')]$$

The Bellman equation is fundamental to RL. Here we only discussed the Bellman equation in relation to MBRL. However, note that the Bellman equation can be used to estimate model-free value as well. For MF value estimation, the equation does not include the transition probabilities among states, p(S' |S, a), but the value of the next state is either based on the actual action taken and direct experience as in Sarsa (on-policy method):

$$Q(s, a) = Q(s, a) + \alpha * [R + \gamma * Q(s', a') - Q(s, a)]$$

or estimated based on the maximum Q-value of the next state regardless of the action taken in Q-learning (off-policy algorithm):

$$Q(s, a) = Q(s, a) + \alpha * [R + \gamma * max(Q(s', a')) - Q(s, a)]$$

So far a candidate model of memory we discussed is model-based RL, with the state transition probability T as a potential model of the relationship among states in memory. In Figure 2 (top left) we considered such a one-step tuple organization (Daw et al. 2011) as one of the hypotheses about how cognitive maps or relational structures may be organized in memory.

We also considered another possibility (Figure 2, top right). Namely, that representations of events and their relationships, or cognitive maps, can be organized beyond one-step relations: as predictive representations of successor states in memory. The RL framework offers a formalization of this perspective in terms of the successor representation (Dayan 1993).

The successor representation (SR) caches multi-step relationships among states. Specifically, it captures how often on average an agent expects to visit a successor state $S'$ starting in state $S$ and moving under policy, $\pi$. Unlike model-based RL, the SR agent doesn't store probabilities of state-action-state transitions. The SR between a given state and a successor state (that can be multiple steps away) is computed by multiplying a discount parameter and the probability of state-action-state transition at every state (see Equation 2).

One way to construct the SR matrix, which we will call $M$ following Dayan's formulation (Dayan 1993), is to derive it analytically from the transition probability of a model-based agent's $T$, where matrix $T$ stores the state-by-state transition probabilities, or the expected transition distribution under the policy. The equation would be as follows:

$$\text{Equation 2. } M \;=\; \sum_{t=0}^{\infty} \gamma^{t} T^{t} \;=\; (I \;-\; \gamma\, T)^{-1}$$

*Where value for state S is computed as:*   $\quad V(S) \;=\; \sum_{S'} M(S,\ S')\, R(S')$

Intuitively, Equation 2 is derived by summing discounted probable states over time, $M = \sum_{t=0}^{\infty} \gamma^{t} T^{t}$, which mathematically converges to $(I \;-\; \gamma\, T)^{-1}$.

However, we do not need to learn $T$ first in order to learn the SR matrix. The SR can be learned via temporal difference learning, in which a row of SR, *M(S)*, is updated as the agent moves from state $S$, as follows:

$$\text{Equation 3. } M(S) \;=\; M(S) \;+\; \alpha\,(\, onehot(S) \;+\; \gamma M(S') \;-\; M(S)\,)$$

This formulation allows for the gradual learning of the successor representation over time (Dayan 1993).

$M$ can be initiated as a matrix of zeros or the identity matrix, when each state could lead to itself. Then, each time the agent visits state *S,* the *Sth* row of M is updated according to Equation 3: with learning rate α, discount parameter γ, and a successor prediction error. At every observation of a state, action, state transition, the *successor prediction error* is the sum of a onehot vector, all zeros except for index S, and a discounted *M(S')* or the S'th row of the successor representation, minus the previous *M(S).* Note that the vector of rewards R is stored separately.

At the moment of decision, unlike model-based RL, the SR agent does not need to roll out every single possible path and decide which one is best. Instead of iterating Equation 1, the expected value can be computed via a linear combination or dot product of $M$ and $R$ (the reward vector). In a sense, SR "pre-bakes" the unrolled combination of probabilities and discount parameters over time, compared to the one-step structure stored by a tabular[1] model-based agent. Therefore, by merely multiplying the reward vector, the SR agent can come up with a decision faster than MBRL.

---

[1] Note that deep MBRL such as DreamerV2 (Hafner et al. 2020) can have representations that go beyond one-step. That said, it's been shown that deep MBRL behavior does not pass tasks requiring the flexibility expected of tabular MBRL (Wan et al. 2022). That said, here we follow a neuroscience tradition that focuses on the tabular notion of MBRL (Daw, Niv, and Dayan 2005; Daw and Dayan 2014; Daw et al. 2011).

A growing number of studies find evidence for the successor representation, and its eigenvectors, in memory and decision-making tasks, alluding to the possibility that it may serve as a general principle for the organization of memories (Gershman et al. 2012; Schapiro et al. 2013; Stachenfeld, Botvinick, and Gershman 2017; I. Momennejad et al. 2017; Garvert, Dolan, and Behrens 04 27, 2017; de Cothi and Barry 2020; Bellmund et al. 2020; Ida Momennejad 2020).

## 2.2. Flexible behavior as a window into latent representations

Imagine you are on the way to work. You notice your favorite food truck is unusually parked a few blocks to the west of the office. During lunch will you default to habit and go three blocks south, where the food truck usually parks, or will you integrate that morning's observation with your knowledge of the city and go west?

An important aspect of planning is flexibility of behavior in response to local changes in the environment that require integrating past memories with new experience. There are at least two kinds of local changes, both of which require flexibility of planning. This could be a change in the location of rewards (like the food truck), or a change in the transition structure (e.g., a road is blocked, requiring a detour, or F train is running on the A train track, etc).

Earlier we discussed RL agents that offer candidate models of memory, namely MBRL and varieties of SR-based algorithms[2]. To address the connection between memory and planning, a reasonable question follows. *Which agent's behavior on planning tasks is more similar to humans'?* One way to discern whether human behavior is more similar to a model-based or SR approach is to design experiments that elicit different behavior in these agents. One such example are experimental paradigms that probe flexibility to local changes in rewards and transition structures, e.g., blocked road, swapped train track (see Section 2.3 for detail).

In the face of a local change (like the food truck example above) model-based RL only needs to have updated the one-step change. If this update was successful while experiencing the local change, MBRL can solve the problem with the same reaction time as planning without such a change. The SR agent, on the other hand, only learns from direct experience (according to Equation 3) and, therefore, can adapt to local changes in rewards (where the structures don't change) but, without using offline replay, is worse at adapting to local changes in transition structures. For instance, it would update the row of the SR associated with the truck but not the row associated with exiting the office. This can be mitigated if the agent could replay the trajectories related to the change it has observed.

---

[2] Recall that model-free agents cannot adapt to *local* changes without re-experiencing the entire trajectory after the change.

This idea that SR could be updated *offline* according to Equation 3 has been introduced in terms of *SR-Dyna* —or *DynaSR (Barnett and Momennejad 2022)*— and hybrid SR-MB (I. Momennejad et al. 2017; Russek et al. 2017). The replay component, Dyna, was proposed in Rich Sutton's proposal of the Dyna architecture, where a model-free agent's policy is updated offline via simulated experience or roll-outs of the model-based T (Sutton 1991).

| | Representation | Computation | Behavior |
|---|---|---|---|
| MF learner | *Q*: Cached value | Retrieve cached value<br>Lowest cost | Habit,<br>Fast |
| MB learner | *R*: Vector of all state rewards<br>*T*: One-step state transitions matrix | Iteratively compute values<br>Highest cost, resource-constrained | Fully flexible,<br>Slow |
| SR learner | *R*: Vector of all state rewards<br>*M*: Multi-step future state occupancy matrix (policy-dependent caching) | Combine cached future occupancies with rewards<br>Intermediate costs | Semi-flexible,<br>Fast |
| Hybrid SR | R & M (as above)<br>SR output combined with T, or update SR with replay (on- or offline) | Combine SR with MB or replay<br>Intermediate costs: mostly SR costs, at times MB or replay costs | Flexible but asymmetric,<br>Fast (mostly) |

**Table 1. Comparison of model representations, value computation, and behavior.** Both the MF cached value and the SR can be learned via simple temporal difference learning during the direct experience of trajectories in the environment. M, the SR or a 'rough' predictive map of each state's successor states; value function Q(s, a) (which maximizes value for optimal policy; compare to V(s), value function for a state under the current policy); R, reward function; T, full single-step transition matrix.

In sum, we have three hypotheses of how latent representations of an environment may be organized when humans engage in flexible planning behavior. The cognitive map, or the relational structure of the environment may be stored as a *1-step map* (MB), a *multi-step map* (successor representation or SR) only learned and updated online, or a multi-step map that can be updated both *online and offline* (SR-Dyna or DynaSR).

The next section dives deeper into tasks that elicit different behavior from these models: the retrospective revaluation of rewards (the structure doesn't change but the rewards change), the retrospective revaluation of transitions (the structure has changed but not the rewards), and empirical results comparing human and agent behavior in these tasks.

## 2.3. Experimental tasks to compare behavior: Model vs. human

Let us return to the central question: How are memory representations structured, and how do these structures connect to prediction and planning? One possibility we discussed is similar to a model-based RL agent that stores one-step transition probability tuples *p(S' | S, a)* and then rolls out possible paths according to Equation 1 to determine the optimal one (Figure 2). Another possibility is that there are multi-step contingencies rather than 1-step transitions. A common version of this is the successor representation or DynaSR, where SR is updated both online and offline.

As noted earlier these models respond differently to tasks with partial changes in rewards and transition structures. Such tasks require the integration of memory and new experience for planning. While model-free RL cannot solve any of the problems (Figure 3), model-based RL can solve both reward revaluation and transition revaluation symmetrically, online SR can only manage reward revaluation, and SR with replay or DynaSR can adapt to both changes. However, the latter needs more computation for transition revaluation, using offline replay to stitch together different pieces of the past to update the present policy.

In a series of experiments, researchers designed a number of simple flexible planning tasks (Figure 3) to measure and compare human and model behavior. Interestingly, human behavioral results reflect an asymmetry (Figure 3, bottom): while humans can solve both tasks, they significantly perform better at retrospective revaluation of rewards compared to transition revaluation. These findings were replicated a number of times in tasks with simple line graph structure and tree structure (I. Momennejad et al. 2017; Russek et al. 2021).

Taken together, human behavioral results are consistent with the DynaSR model, which predicts both asymmetric accuracies on reward and transition revaluation as well as longer reaction times for transition revaluation. Another behavioral result to note is that this asymmetry is not the result of a speed accuracy trade-off, that is worse performance is not accompanied by faster reaction times (RTs). In fact, on the contrary, human reaction times are significantly longer for solving transition revaluation (on which human performance was less accurate) compared to reward revaluation (Figure 3, bottom right).

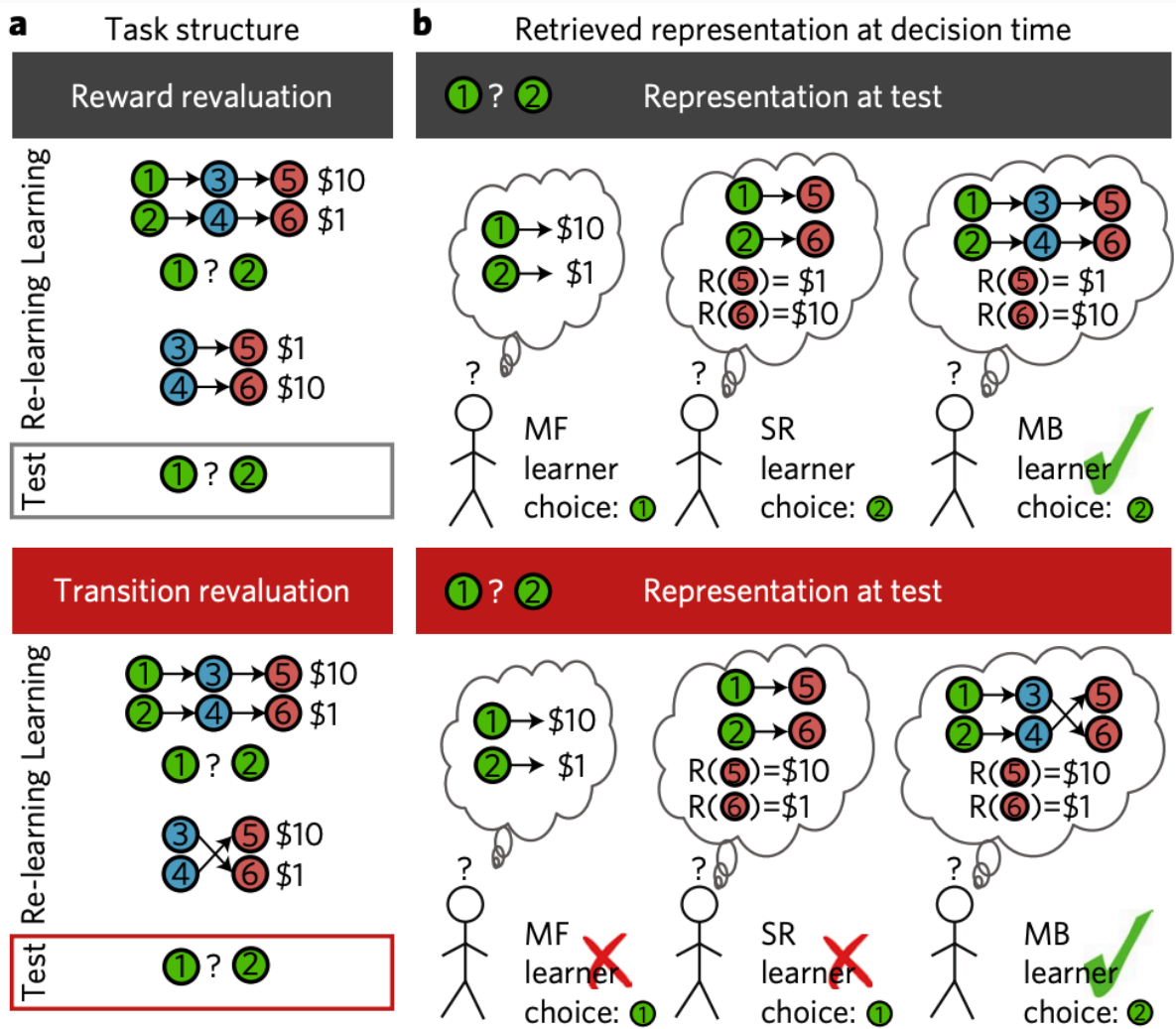

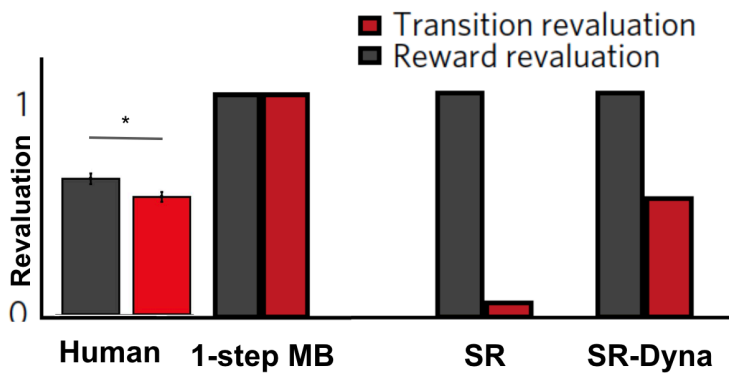

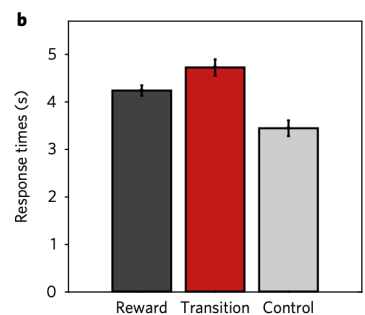

**Figure 3. Experimental design, hypothesized representations, human and model behavior.** (Top) The design of reward and transition revaluation and a schematic illustration of how each of the three RL agents discussed earlier would solve the problem. (Bottom) Model-free agents cannot solve either, model-based is expected to solve both equally well, a successor representation learner without replay is expected solve reward revelation but fail at transition revaluation, and a hybrid agents learning SR both online and offline via replay (SR Dyna or DynaSR) is expected to learn both but be better at reward revaluation. Human behavior in terms of both accuracy and reaction times is more consistent with DynaSR agents.

Accounting for both differences in accuracies and reaction times is important for computational models of brain function. For instance, a recent study considered the hypothesis that a probabilistic successor representation may capture the same behavioral results. However, while the probabilistic approach captures the asymmetry in accuracies in its present format it cannot trivially account for reaction time differences observed in humans, without ad hoc additions and assumptions (J. P. Geerts, Stachenfeld, and Burgess 2019).

Methodologically, the importance of explaining various dimensions of behavior (e.g., accuracy and RT) in computational models of cognitive capacities and brain function cannot be overstated. This is especially important in the field of neuroAI and building human-like agents (Ida Momennejad 2023).

## 2.4. Prioritized replay during offline learning

We discussed the importance of offline learning via replay as a key component of DynaSR, the model that best captured human behavior in tasks that required integrating memory and planning (Figure 3). However, in real life problem settings there are far too many memories to replay and reactivate, such that were brains to replay them at random, noticing the most relevant memories and stitching them together could become a matter of luck, or even intractable. So how can our models of brains account for choosing the most relevant memories to replay?

One possibility is that the model that best captures behavioral and neural findings *prioritizes* which memories to replay according to the amount of surprise associated with them. This notion of surprise can be captured in terms of prediction errors, computed as the difference between what the agent expects and what it observes, which can be signed (positive or negative prediction error) or unsigned (simply a surprising event). This idea has been used in reinforcement learning since the 90s (A. W. Moore and Atkeson 1993) and still echoes in contemporary RL and deep learning algorithms as much as in psychology and neuroscience (Sutton et al. 2012; Schaul et al. 2015; Mattar and Daw 2018; Rouhani and Niv 2021).

While it is more common to consider reward prediction errors in the context of memory prioritization, it is also possible to consider successor prediction errors (PE) as discussed earlier in relation to Equation 3. A recent modeling paper proposed PARSR, Priority Adjusted Replay for Successor Representations, offering variants of DynaSR where the replay priority can be set to reflect *reward PE* or *successor PE* (Barnett and Momennejad 2022). The study found differences in the representations learned by the different prioritization approaches, which may better serve different types of tasks. Future research is required in this area to better elucidate the usefulness of these different approaches.

### 3. Neural representations: Model vs. Human

### 3.1. Evidence for predictive representations

We discussed behavioral evidence for the hypothesis that cognitive maps may be organized similarly to the successor representation: predictive representations of expected visitations to successor states. SR has a number of properties that lend well to a number of testable neural predictions. For instance, the rows of SR capture the future, or the successors of each state, and the columns of which capture the past, or the predecessors of each state (Figure 4).

Importantly, it has also been shown that the columns of SR resemble the representation of place fields in the hippocampus, and explain observed phenomena including backwards expansion on tracks and elongation near walls (Stachenfeld, Botvinick, and Gershman 2017; George et al. 2022). Note that policy-dependent (or path-dependent) SR representations predict that the brain represents distances not in Euclidean terms, but in a path-dependent fashion, see Figure 4 (Russek et al. 2017; Ida Momennejad 2020).

This is in line with a number of observations in rodent and human neuroscience. Mehta and colleagues have shown that place fields become more predictive with experience, i.e., they expand along the direction of the path experienced (Mayank R. Mehta, Barnes, and McNaughton 1997; M. R. Mehta, Quirk, and Wilson 2000), see the rightmost representation in Figure 4. Moreover, distance to goals has been shown to be path dependent and not Euclidean in human medial temporal lobe (Spiers and Maguire 2007; L. R. Howard et al. 2014). This is in line with a multi-scale SR model discussed later, the derivative of which can capture distance among states (I. Momennejad and Howard 2018), see Section 3.4 for more detail.

## Multi-step representations of relational structures

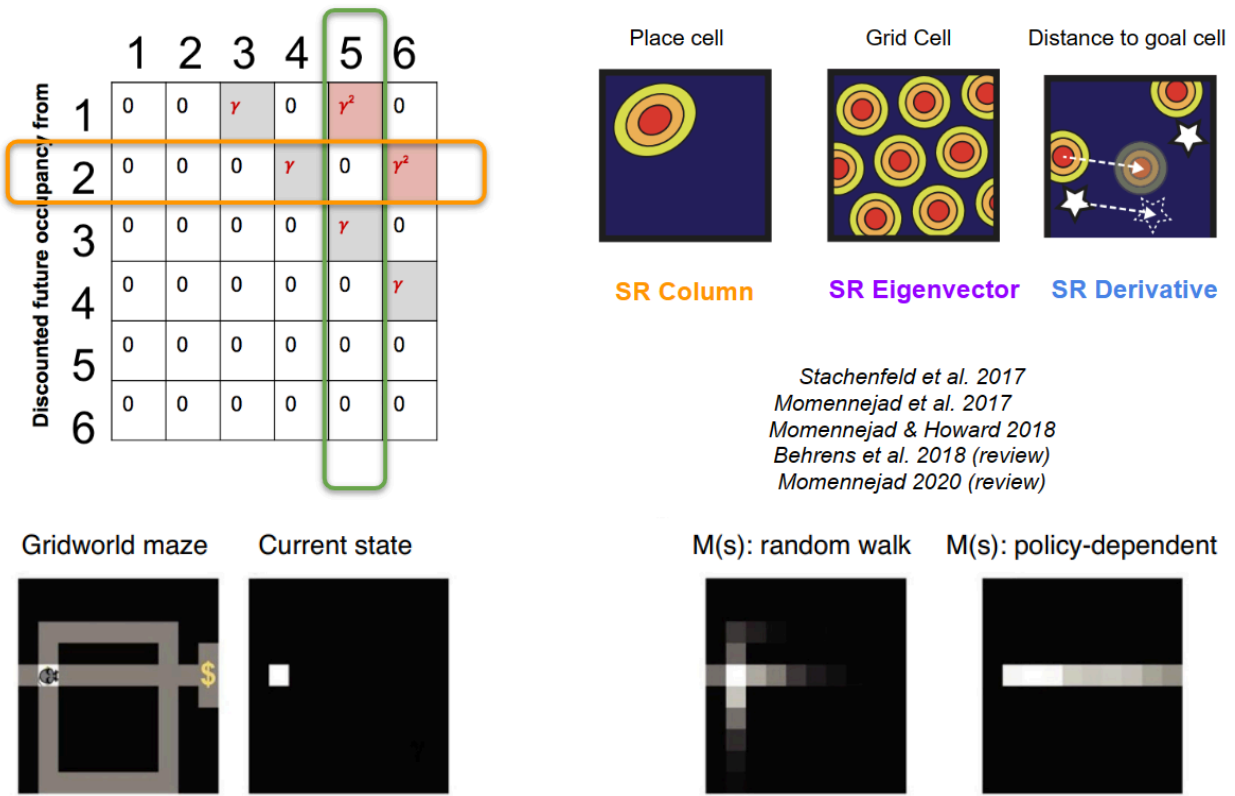

**FIgure 4. Successor representation as a model of predictive cognitive maps.** (Top Left) A successor representation matrix for the graph structure of the task in Figure 3 is displayed. The rows correspond to the successor states that are one or more steps away, or the future (e.g., the successors of state 2 have non-zero successor representations in row 2). The columns represent the predecessors of a state (or the past). NB: the diagonal can also be 1, assuming there's a possibility of going to a state from that state. (Top Right) Evidence suggests that columns of SR simulate place fields in the hippocampus, while the eigenvector of SR simulates grid fields, and the derivative of multiscale SR estimates the distance among states. These findings make SR, a path or policy-dependent representation, a viable computational candidate for capturing memory and space in the medial temporal lobe (MTL). (Bottom) Consider a rodent in the gridworld maze. Its current location, its successor representation according to a random walk

(independent of the path it takes) and according to the policy or path it takes are displayed. Place fields become predictive of the path with experience .

A human fMRI study (Russek et al. 2021) conducted the flexible planning experiment discussed Section 2.3 (I. Momennejad et al. 2017) in the scanner. The study focused on transition revaluation (Figure 3), investigating whether fMRI evidence for outdated successor representations (i.e., persisting representational similarity of a starting state to a further state that is no longer a successor) could predict behavioral errors in transition revaluation. They found that indeed, in regions associated with navigation and planning, representational similarity to outdated successors predicted behavioral errors (Russek et al. 2021).

If the SR captures the structure of representations that underlie navigation, some further behavioral predictions could be made. A recent paper specifically investigated how deforming the cognitive map may distort memories and showed how a model with SR and its eigenvectors can explain the phenomenon (Bellmund et al. 2020). The study used an innovative design, in which the structure of the environment goes from a rectangle to a trapezoid, leading to distortions in behavioral judgments such as distance.

Another study proposed a model of how environmental manipulations may impact predictive representations in the medial temporal lobe. They proposed a model where the SR is learned from a basis set of boundary vector cells (BVCs), and showed that the model captures place cell firing in terms of successor features, while grid cells represent a low-dimensional representation of these successor features (de Cothi and Barry 2020). They test the predictions of the model against environmental manipulations such as dimensional stretches, barrier insertions, and the influence of a room's geometry on the representation of space (related to the behavioral study mentioned above).

## 3.2. Connecting the Past and the Future: Multiscale Predictive representations

Let us return to the example of planning a flight, and moving between coarse and more detailed scales during planning. In the models described earlier, the scale or horizon of abstraction of a given successor representation is determined by the discount parameter, $\gamma$. However, in the agents we discussed so far only one discount parameter was used to learn the relations structure of the environment. Yet, various scales of abstraction might be more appropriate for different planning problems. How does the brain, or an RL agent, handle planning seamlessly at different scales of granularity?

If we take the successor representation approach, one hypothesis is that the brain or the model simply stores multiple successor representations with different values of the discount parameter γ. In other words, the representation of the environment stores multi-step dependencies according to different scales or discount parameters, leading to a multi-scale set of successor representations.

If such a model captures multi-scale representations in the brain, a number of testable empirical predictions follow. One prediction is that during navigation of long distances, representations in different parts of the brain should reveal sensitivity to different horizons. If not, all brain regions should show the same level of granularity.

In order to test this hypothesis, it is important to go beyond small-scale laboratory studies of how relational knowledge enables inference and planning in few step controlled designs. Such a study needs to be feasible for neural measurements, e.g., human neuroimaging. Virtual reality-like navigation of long distances inside an fMRI scanner offers a sweet spot between laboratory and life-scale planning inside a scanner, offering a window into studying how people use stored knowledge in continuously unfolding navigation, e.g., walking long distances in a city.

One study used an existing fMRI dataset of virtual navigation of realistic distances (of up to kilometers) in the city of Toronto (Brunec et al. 2018). They hypothesized that there are predictive representations organized at multiple scales along posterior-anterior prefrontal and hippocampal hierarchies, and that they guide naturalistic VR navigation (Brunec and Momennejad 2021). To test this hypothesis, the study conducted model-based representational similarity analyses of neuroimaging data measured during navigation of realistically long paths in VR. They tested the pattern similarity of each point along a given path to a weighted sum of its successor points within different predictive horizons (Figure 5). They found that the anterior prefrontal cortex (PFC) showed the largest predictive horizons (up to a kilometer), posterior hippocampus the smallest (about 25 meters), with the anterior hippocampus and orbitofrontal regions in between (a hundred to three hundred meters). These findings offer novel insights into how multi-scale cognitive maps can connect memory and prediction to support hierarchical planning (Figure 5, bottom).

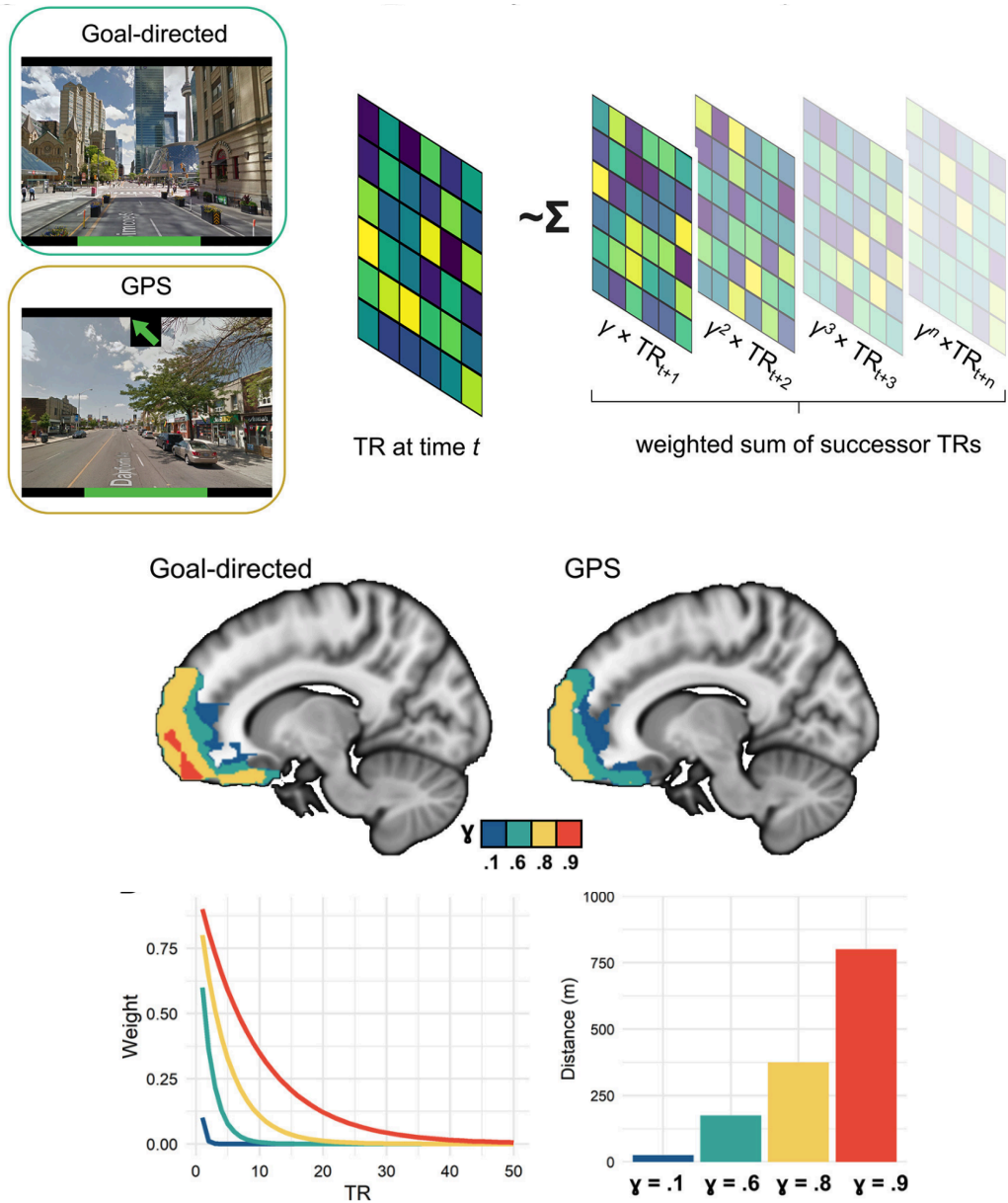

**Figure 5. Multiscale successor representations.** (Top) The virtual Toronto navigation experiment was conducted in fMRI with two conditions: goal-directed, in which participants navigated to a known goal with no guidance and based on their memory, and GPS, in which they did not know the goal and followed a GPS arrow in an unfamiliar part of the city. The similarity of fMRI representations was measured between every point on the trajectory and the weighted sum of its successor states on the taken path using different discount parameters. (Bottom)

In the prefrontal cortex, the largest horizon or scale was only observed in the goal-directed condition (in the anterior PFC, up to about 1 kilometer). Smaller horizons, which could have been seen in the VR image, were observed in the GPS condition as well.

A number of recent studies consistent with the multi-scale view here are noteworthy. A recent study simultaneously recorded hippocampal-prefrontal ensembles and investigated how rats generalize navigational rules across different environments. They showed that the hippocampus represented the specificity of separate environments, while the prefrontal cortex representations generalized across environments (Tang, Shin, and Jadhav 2023). A human study investigated abstract representations of a story in human brains as participants listened to the story (Owen, Chang, and Manning 2021). They found that the anterior hippocampus and prefrontal cortex had more information about paragraphs and longer scale themes in the intact stories, than words and jumbled versions of the story.

While some studies propose emergent self-scaling in entorhinal grid fields (Fiete, Khona, and Chandra 2023), other studies explored models of mechanisms underlying multi-scale memory retrieval. They suggest that inhibitory diversity can increase the range of memory retrieval and change how the network of memories becomes activated (Burns, Haga 芳賀 達也, and Fukai 深井朋樹 2022). Moreover, recent bat studies have reported multiscale representation of larger spatial environments in the bat hippocampus (Eliav et al. 2021). Taken together, this body of studies offers evidence in support of multi-scale memory representations across species and using a diverse set of imaging, empirical, and theoretical methods.

### 3.3. Neural evidence for offline replay

Earlier we discussed the importance of offline replay in the models that best capture human behavior in flexible planning (Figure 3). Let us consider neural evidence that offline replay of past states contributes to updating a planning policy.

A human neuroimaging study used a behavioral design with reward revaluation (where the rewards change but the may of the environment doesn't change, Figure 3) and a control condition (I. Momennejad et al. 2018), and interleaved three 30-second rest periods during the re-learning phase (in which participants no longer visited the starting states). The study showed that while offline replay of earlier states that had not been visited in a while (Figure 3, state 1) was correlated with changing planning behavior in the test phase, there was no correlation between offline replay and behavior in the control condition (where nothing changes in the world).

Notably, the study did not use hippocampal replay, which is too fast for fMRI and hippocampal fMRI signals are often noisier than some other regions[3]. That said, the reactivation of cortical patterns related to an earlier state. This memory reactivation follows the expected patterns hypothesized by replay that prioritizes surprise. Moreover, the study showed that the extent to which the brain was sensitive to unsigned prediction error (or surprise) during the learning phase, could predict how much the earlier states would be replayed during rest, and in turn, how much this offline replay would correlate with subsequent changes in revaluation behavior during the test phase.

Over the past decade a number of studies have investigated the role of replay in behavior across different species and using different imaging modalities. A series of human fMRI studies have shown evidence for offline replay in non-spatial sequential tasks across the hippocampus, visual cortex, and the orbitofrontal cortex (Schuck and Niv 2019; Wittkuhn and Schuck 2021; Schuck et al. 2016; Wittkuhn et al. 2021).

A series of rodent studies suggest that forward and backward or reverse replay may differentially contribute to planning and updating a plan following prediction errors, forming a cognitive map by capturing the topological map of the environment (Brad E. Pfeiffer and Foster 2013; B. E. Pfeiffer and Foster 2015b; Foster and Wilson 2006; Wu and Foster 2014; Foster 2017; Ólafsdóttir et al. 2015; B. E. Pfeiffer and Foster 2015a; Widloski and Foster 2022). A recent study suggests that hippocampal replay appears after a single experience but episodic details may emerge with more experience  (Berners-Lee et al. 2022).

A number of other studies have established the role of prediction errors and surprise (unsigned prediction errors) in memory in both laboratory tasks (Rouhani and Niv 2021) as well as natural and long-term life events (Rouhani, Stanley, et al. 2023). Other research has highlighted the role of uncertainty and prediction errors in offline learning and memory (Rouhani, Niv, et al. 2023; Schapiro et al. 2018). More research is needed to better understand whether there are varieties of prioritization of memory reactivation and replay depending on the task demands, time pressure, and memory resources (Barnett and Momennejad 2022). It is also possible that such task-specific prioritization is meta-learned during the life-long learning of processes that are optimal for different classes of generalized problems (Jane X. Wang et al. 2018; J. X. Wang et al. 2016; Botvinick et al. 2019).

---

[3] N.B. Some suggest that specific experimental designs can overcome this problem of fast hippocampal replay (Schuck and Niv 2019). However, it is the author's opinion that the feasibility and reliability of this approach for other study designs that may require longer learning and relearning phases as well as rest periods remains to be established.

One area that we did not address in depth within this chapter is the role of sleep in replay and the reorganization of memory. Consistent with models, behavioral, and neural evidence discussed in sections 2 and 3, cognitive neuroscience research suggests that sleep improves memory representations in humans (Coutanche et al. 2013). This body of work lends further evidence to the idea that replay and reorganization of memories during sleep may support generalization, semantic memory, and flexible behavior (Tandoc et al. 2021; Schapiro et al. 2018; Poe, Walsh, and Bjorness 2010; Schapiro, McDevitt, et al. 2017).

### 3.4. The successor representation and episodic memory

So far we have only discussed hippocampal function in terms of multi-step predictive representations and temporal abstraction. Neuroscience has provided mechanisms and evidence for how synaptic plasticity in the hippocampus leads to sequence learning and spatial maps (M. R. Mehta, Lee, and Wilson 2002; Mayank R. Mehta 2015) and at different scales (J. J. Moore et al. 2021). However, beyond capturing the relational structure of events, the hippocampus is also involved in episodic memory (Tulving and Markowitsch 1998), or memory of individual events. Are both these two seemingly different functions supported by successor representations?

A computational study has proposed a link between the successor representation, eligibility traces, and the temporal context model (TCM) of episodic memory (Gershman et al. 2012). According to TCM, as we navigate the world the brain's mental context dynamically changes over time in response to both internal and external events, and anything we store in memory is bound with the temporal context during encoding. Therefore, the temporal context can serve as a cue for retrieval, explaining the way in which memories encoded close in time (i.e., have a shared temporal context) are recalled together as a cluster. TCM offers quantitative explanations of the recency effect (recent memories are easier to recall) and the contiguity effect (adjacent items are easier to recall, especially a successor adjacent item), and related medal temporal lobe function (M. W. Howard and Kahana 2002; Polyn, Norman, and Kahana 2009-1; M. W. Howard et al. 2005-1; M. W. Howard, Youker, and Venkatadass 2008-2; Mau et al. 2018).

The computational study suggested that the Temporal Context Model (TCM) is estimating the SR using temporal difference learning–as described earlier (similar to Equation 3) (Gershman et al. 2012). They suggest that the successor prediction is formed through recurrent dynamics in the CA3 subfield of the hippocampus, which is then compared to sensory input arriving at CA1 directly from the entorhinal cortex (EC), thus, CA1 computes successor prediction error, consistent with the reported novelty (mismatch) signal in this region (Lisman and Otmakhova 2001; Kumaran and Maguire 2005, 2007).

A more recent study suggests that error-driven temporal difference learning may not be implemented in hippocampal networks (George et al. 2022). Rather, they suggest that spike-timing dependent plasticity (STDP), a form of Hebbian learning, may rapidly learn an approximation of SR from "theta sweeps", or temporally compressed trajectories. The model uses spiking neurons modulated by theta-band oscillations to capture SR-related phenomena (e.g., backwards expansion on a 1D track and elongation near walls in 2D (Stachenfeld, Botvinick, and Gershman 2017)), as well as multi-scale place field sizes along the dorsal-ventral axis of the rodent hippocampus. The authors suggest that such topological ordering is necessary to prevent larger place fields from mixing up the scales (George et al. 2022).

Another study looked at the mathematical relationship between the successor representation and another prominent model of remembering multiscale memories, which models long term memory using the Laplace transform of past events (I. Momennejad and Howard 2018). The study noted that SR with a single scale often discards information about the sequential order of states and the distance between them. Given these are task-relevant in many navigation tasks in animals and artificial agents, the paper suggested that establishing the plausibility of SR as an organizing principle for the past and future requires an approach that can reconstruct the sequence.

The authors proposed that operations on an ensemble of SRs with multiple scales can reconstruct both the sequence of future states and estimate the distance to goal (Figure 5, bottom left). The computation needed was simply to compute the derivative of SR between a given $S$ and $S'$, which can be computed linearly. They showed that a multi-scale SR ensemble is mathematically equivalent to the Laplace transform of future states, and the inverse of this Laplace transform is a biologically plausible linear estimation of the derivative. This suggests the possibility that multi-scale SR and its derivative could lead to a common principle for how the medial temporal lobe supports both map-based and vector-based navigation.

In short, the study showed that multi-scale successor representations and their derivatives (I. Momennejad and Howard 2018) are mathematically equivalent to a prominent computational model of memory (Shankar and Howard 2012; Shankar, Singh, and Howard 2016), in which the inverse laplace transform of a sequence of past events corresponds to distance of the past memory to the present. Notably, the multiscale SR approach and its derivative could explain distance to goal cells observed in bat hippocampal data (Sarel et al. 2017). Note that not only can individual state trajectories and their distance be recovered from multi-scale SR, but since the Laplace transform and its inverse had been previously proposed to underlie remembering past events in a scale-invariant fashion (Shankar and Howard 2012; M. W. Howard et al. 2014; Shankar, Singh, and Howard 2016), these findings add credence to the idea of multiscale SR as a principle for memory organization in the medial temporal lobe.

### 3.5. Predictive representations and complementary learning systems

A related computational view suggests the complementary learning systems in the hippocampus: that it serves both learning distinct separate episodes as well as general statistical commonalities among the episodes (Schapiro, Turk-Browne, et al. 2017). They used a neural network rather than the RL framework to ask how the hippocampus handles both statistical learning (which requires detecting commonalities) and memorization of individual episodes (Schapiro, Turk-Browne, et al. 2017). The neural network model, with hippocampus-like connectivity and subfields, was trained on sequences with temporal regularities, similar to the stimuli in statistical learning experiments. The results suggest that the pathway connecting the entorhinal cortex directly to CA1 may support statistical learning (monosynaptic pathway), while the pathway connecting entorhinal cortex to CA1 through dentate gyrus and CA3 (trisynaptic pathway) learns individual episodes.

According to this view, associative reactivation through recurrence may give rise to the representations of statistical regularities, and different anatomical pathways may mediate the trade-off between learning episodes and associative functions of the hippocampus. This is also in line with a study suggesting distinct roles for dorsal CA3 and CA1 in memory for sequential nonspatial events (Farovik, Dupont, and Eichenbaum 2010). A more recent human neuroimaging study used an associative learning paradigm to test both SR and the complementary learning systems predictions, and found intriguing convergence (Pudhiyidath et al. 2022). Future research can further illuminate this link between SR and complementary learning and memory systems.

Consistently, neuroimaging studies suggest a complementary role for CA1 and CA3 in episodic memory and context. One study showed that CA1 represented objects that shared an episodic context as more similar to each other, while CA3 differentiated between objects encountered in the same episodic context (Dimsdale-Zucker et al. 2018). The authors suggest that the complementary nature of CA1 and CA3 captures how we parse experiences into cohesive episodes while retaining the specific details. A consistent study showed that damage to CA3 can disrupt both recent and distant episodic memories (Miller et al. 2020).

Taken together, these studies suggest that hippocampal subfields may contribute to both associative and episodic learning through complementary circuitry, functions, and interactions. Moreover, recent work connects the statistical learning of successor representation to on-task replay (Wittkuhn, Krippner, and Schuck 2022), providing further credence to the importance of the predictive structure of memory in our understanding of behavioral and neural phenomena surrounding memory, replay, and planning.

## 3. 6. Predictive representations, hierarchical planning, and deep RL

Most real world navigation and planning problems involve hierarchical problem solving. Studies of taxi drivers and indigenous navigators offer evidence for the explicit use of hierarchical planning in humans in the real world (Spiers and Maguire 2008; Fernandez-Velasco and Spiers 2024). Behavioral and neural signatures of hierarchical planning and navigation, as well as their computational models, are active areas of research in contemporary computer science and cognitive neuroscience.

Hierarchical RL (or HRL) focuses on representation learning mechanisms that empower an agent to explore an environment and learn options, which are abstractions of policies or various sequences of state-action-state sequences that functionally achieve the same goal (Sutton, Precup, and Singh 1999; Dietterich n.d.; Stolle and Precup 2002; Barto and Mahadevan 2003; Bacon, Harb, and Precup 2016; Botvinick and Weinstein 2014; Xia and Collins 2020). The HRL and options framework allow for more efficient exploration and representations for goal-directed navigation and problem solving.

Moreover, neuroscientific studies of hierarchical planning and reasoning suggest a key role for the prefrontal cortex, e.g., in motivating functional hierarchies in cognitive control (Egner 2009) and simulating the future (Javadi et al. 2017). Patient studies show that damage to the human anterior prefrontal cortex leads to impairments in multi-tasking, or completing a sequence of errands in order (P. W. Burgess 2000; P. W. Burgess et al. 2000; Roca et al. 2011). A number of studies combining human fMRI and machine learning approaches have established a role for the anterior prefrontal cortex in prospective memory, or remembering to execute a long-term intention later while we are busy doing something else now (Okuda et al. 2007; Paul W. Burgess et al. 2008; I. Momennejad and Haynes 2012; Ida Momennejad and Haynes 2013; Haynes et al. 2015). Primate neural recordings also suggest a role in hierarchical reasoning by the frontal cortex (Sarafyazd and Jazayeri 2019).

Recent work in deep learning used inspiration from human studies of collective memory and collective cognition (Coman et al. 2016; Hirst, Yamashiro, and Coman 2018; Ida Momennejad, Duker, and Coman 2019; Ida Momennejad 2022) to propose a multi-agent approach to hierarchical problem solving using deep RL agents (Nisioti et al. 2022). The study designed a set of hierarchical problems with various levels of branching and breadth, and connected a network of 10 deep Q-learning networks (DQNs) with different connectivity graphs, and had each multi-agent collective of agents solve the problems by exploring individually and sharing their replay content with each other (Nisioti et al. 2022). The results revealed that it was not an all-to-all connectivity nor individual agents that could optimally solve all problems. The more difficult problems could only be solved by networks with dynamic connectivity: where agents first communicated in smaller clusters and then changed the connectivity to communicate more broadly, and repeated this dynamic connectivity.

This multi-agent finding is potentially useful in understanding how the prefrontal cortex flexibly coordinates communication among other brain regions to solve different tasks such as hierarchical problems. While some tasks may be better served by dense connectivity among certain brain regions, others may require selective or dynamic connecting of some connections, and the PFC may need to metalearn optimal connectivity patterns for each problem category over the course of continual and lifelong learning (Lopez-Paz and Ranzato 2017; Parisi et al. 2018). Future research could shed light on such a multi-agent communication hypothesis about the role of the PFC in flexible task switching and hierarchical problem solving.

Let us return to multiscale predictive representations. Over the past decade, a series of RL studies have linked the options and hierarchical RL framework to various forms of the successor representation. A number of RL studies have focused on the eigenvectors of the graph laplacian of the environment's structure to derive options and subgoals in HRL (Jr and Morley 1985; Gutman 2003; Sprekeler 2011; Machado, Bellemare, and Bowling 2017; Klissarov and Machado 2023), as well as eigen-options, or generalized option components, the linear combination of which can provide various policies in an environment (Sherstan, Machado, and Pilarski 2018; Machado et al. 2017; Machado, Bellemare, and Bowling 2020). Notably, SR is comparable to the inverse of the graph laplacian. Thus, a number of studies investigated how successor feature learning and deep successor representation models acquire representational underpinning of HRL.

A standing challenge in this area and for future studies involves the observed dichotomy between the learning rules in deep SR, or deep successor feature agents, and the emergent structure of learned representations. While the learning rules follow SR, sometimes the similarity structure of representations do not reflect what would be expected to a tabular successor representation approach. Future studies are required to better understand efficient and optimal learning and use of SR and deep SR in hierarchical learning and problem solving.

Future research is required to better understand more prefrontal cortical centered functions such as task sets and schema (Farzanfar et al. 2023; Masís-Obando, Norman, and Baldassano 2022; Preston and Eichenbaum 2013; Graziano and Webb 2015; Gilboa and Marlatte 2017; McKenzie et al. 2014; Gupta et al. 2012), compositional tasks, and decomposing value and policy with different basis sets (Whittington et al. 2019).

## 4. Evaluating generative AI's planning behavior and underlying representations

### 4.1. Evaluating navigation behavior in xbox games

Contemporary generative AI is ubiquitous, from search engines to medical assistants, office copilots, and game agents, various dimensions of the future of life and work are tied to it (Lee, Goldberg, and Kohane 2023). However, various capacities discussed here related to planning, long-term memory with efficient retrieval, and navigation remain unresolved challenges for contemporary generative AI.

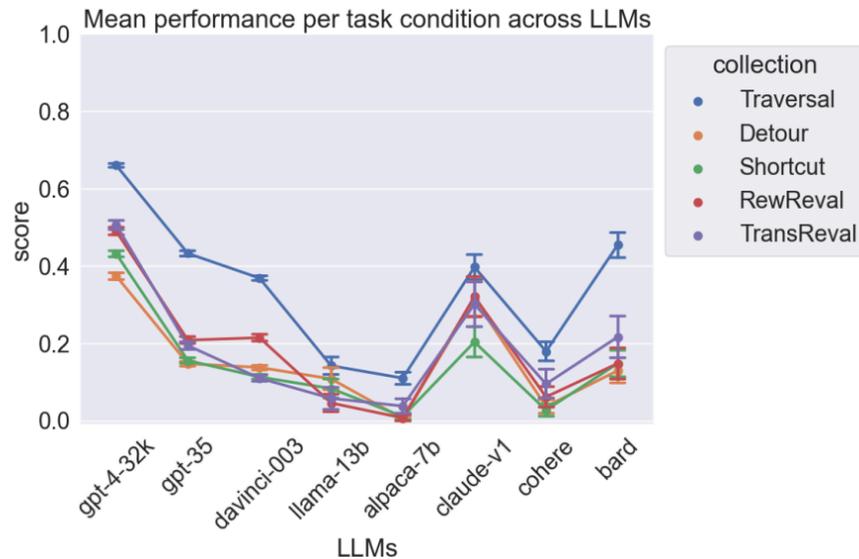

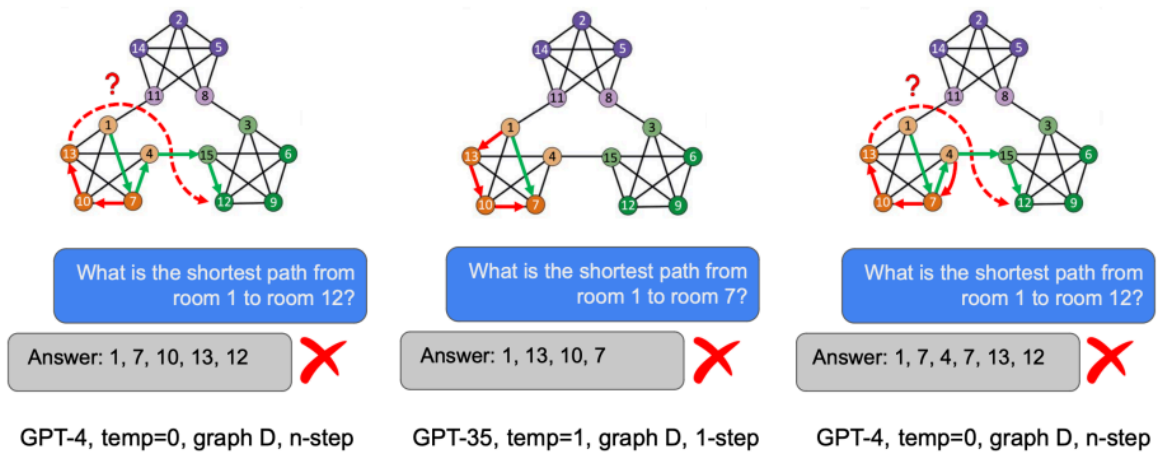

**Figure 6. Evaluating planning and cognitive maps in Large language models.** (Top) The performance of 8 LLMs on tasks related to flexible planning. (Bottom) Three major failure modes of LLMs include hallucinating edges that do not exist (there is no direct path between 13 and 12), taking a long path when a 1-step path is available, and falling in loops (I. Momennejad et al. 2023). These failure modes do not support the idea that LLMs including GPT-4 have emergent cognitive maps or planning ability.

It is the author's belief that cognitive science and computational neuroscience inspired methods and theories can enhance both the evaluation of generative AI systems and inspire architectures and solutions that enhance their abilities. This may herald a new dawn for the application of cognitive neuroscience approaches in AI.

While a detailed survey remains outside the scope of the present chapter, it is noteworthy that the classic computer science benchmarks have proven to not be sufficient for the evaluation of memory, generalization, planning, and navigation, and methods from cognitive and neural science prove highly relevant. For instance, two studies compared the behavior of human-like agents with human behavior when navigating in the same VR game environment: Bleeding Edge (Devlin et al. 2021; Zuniga et al. 2022). They showed that while both agents had similarly high performance on benchmarks such as reward and steps to completion, neither of them passed the Navigation Turing Test (HNTT).

That is, human judges observed side by side videos of an avatar navigating the game, and could significantly tell in which videos the avatar's navigation was controlled by agents and in which by humans (Devlin et al. 2021; Zuniga et al. 2022). Notably, a second study use six different artificial judges to determine if they can judge human-like navigation in the game, and found that while they reached high performance at detecting human play, when comparing videos played by two agents side by side, they could not distinguish or rank which of the two agents would be judged as navigating more human-like by human judges (Devlin et al. 2021).

The significance of these findings are twofold: first, benchmarks are not enough to capture human-like behavior–see this rubric for human-like agents and neuroAI for more detail (Ida Momennejad 2023); and second, judging which of two agents display human-like navigation was not trivial for existing imitation learning and deep learning models. A deeper understanding of both human navigation behavior and metrics for its understanding are required to address these challenges.

## 4.2. Can large language models plan?

Another intersection of the research discussed in this chapter with generative AI is in the evaluation of planning and navigation behavior in large language models (LLMs) such as GPT-4, Bard, or Llama. A recent study (I. Momennejad et al. 2023) turned the design of a number of studies mentioned in this paper (including Figure 3) into prompts, and tested eight LLM's behavior on variety of planning related tasks, such as reward or goal directed planning, finding shortest paths during multi-step traversal, reward revaluation, transition revaluation, shortcut, and detour tasks (Figure 6).

The paper proposed CogEval (I. Momennejad et al. 2023), a cognitive-science inspired protocol for the systematic evaluation of cognitive capacities (such as planning and cognitive maps) in LLMs. From Tolman to the present day, flexible planning behavior in humans and animals has been studied in terms of robustness to various local changes in the environment that have implications for the global policies, which is the same approach used by the researchers to study the ability to extract and use cognitive maps for planning in eight LLMs.

The study found that taken together, none of the eight large language models tested showed consistent flexible behavior across the tasks. Notably, while larger models (especially GPT-4) showed apparent success on some of the simpler linear planning tasks, a number of failure modes suggest that even GPT-4 does not have strong emergent planning capacities. These failure modes included hallucinating edges and paths that didn't exist, falling into loops, and taking many steps of unnecessary moves even when the LLM merely needed to traverse a one-step path. Together, planning performance and failures in LLMs including GPT-4 do not support the idea of emergent planning or cognitive map capacity.

Follow up research (Webb et al. 2023) inspired by cognitive computational neuroscience methods, improved LLM-based planning by creating iterative prompts that function like different components of the prefrontal cortex (PFC). This led to a modular black-box architecture, where GPT-4 was prompted to play the role of different PFC regions to solve the problem in an iterative fashion (Webb et al. 2023). The results show improvement in hallucinations as well as planning ability in both graph traversal and Tower of Hanoi tasks. Future research can improve on this modular approach by creating multi-scale modules that can identify subgoals.

## Summary

How do memories simultaneously capture details about the past, while enabling generalization, prediction, and planning? Here we reviewed evidence for the hypothesis that the answer involves how memories are organized in the brain, namely, as multi-scale predictive representations. We reviewed computational, behavioral, and neural evidence for this hypothesis, and showed that it is compatible with complementary memory systems, in which different subregions and gradients of the brain's memory systems collaborate to support episodic details on the one hand, and abstraction on the other: the past and the future.

We discussed reinforcement learning as one of the computational frameworks for understanding how predictive representations may be organized in memory. The successor representation (SR) was discussed as a candidate principle for generalization in computational accounts of memory, and the structure of predictive representations in the hippocampus and the prefrontal cortex.

We discussed how SR is learned when an agent navigates a sequence of states: the SR stores how often, on average, each upcoming state is expected to be visited. This is different from other RL accounts such as model-based RL that learn only 1-step associations, in that SR learns expected this future visitation for states that are multi-steps away. A discount or scale parameter determines how many steps into the future SR's generalizations reach, which in turn enables rapid value computation, subgoal discovery (e.g., via SR's eigenvectors), and flexible decision-making in larger decision trees.

When an agent, or the brain, learns multiple SRs with different discount or scale parameters, we have a multiscale set of predictive representations, the rows of which capture the future, or the successors of each state, and the columns of which capture the past, or the predecessors of each state (Figure 4). It has been shown that the successor representation (SR) offers a candidate principle for generalization in reinforcement learning (Dayan, 1993; Momennejad, Russek, et al., 2017; Russek, Momennejad, Botvinick, Gershman, & Daw, 2017) and computational accounts of episodic memory and temporal context (Gershman, Moore, Todd, Norman, & Sederberg, 2012), with implications for neural representations in the medial temporal lobe (Stachenfeld, Botvinick, & Gershman, 2017) and the midbrain dopamine system (Gardner, Schoenbaum, & Gershman, 2018).

The central question was echoed in our discussion of complementary learning systems in the medial temporal lobe (Schapiro et al. 2013). The view suggests that hippocampal subfields contribute to both associative and statistical learning, as well as learning of distinct episodes. We discussed how predictive representations and the successor representation could be used to model this theory as well.

We also discussed how a combined computational, behavioral, and neural evaluation methodology can be used to evaluate and improve generative AI. We discussed an example of a deep RL agent navigating an xbox game, showing that merely beating certain benchmark metrics such as total rewards or steps to goal are not sufficient to guarantee human-like navigation behavior in the xbox game. Possible solutions may indeed require a more brain-like approach to the architecture of the agents. We also summarized research evaluating and improving planning behavior in large language models, using similar empirical paradigms to those summarized in this chapter.

Taken together, evidence reviewed here supports the idea that memories and cognitive maps may be structured in terms of multiscale predictive representations along hippocampal and prefrontal hierarchies, supporting flexible behavior in humans, rodents, bats, and agents. Future research can shed further light on the processes that use these representations for further abstraction, schema, and task-based generalization mediated by these regions.

**Acknowledgements**

The author is profoundly grateful to Kim Stachenfeld for comments on the manuscript, and Lynn Nadel and Sara Aronowitz for their kindness, resourcefulness, and patience in both organizing workshops as well as this book.